\documentclass[11pt]{article}
\usepackage[margin=1in]{geometry}
\usepackage[T1]{fontenc}
\usepackage[utf8]{inputenc}
\usepackage{lmodern}
\usepackage{amsmath,amssymb,amsthm}
\usepackage{booktabs,longtable,array,tabularx}
\usepackage{tikz}
\usetikzlibrary{arrows.meta,positioning,fit,calc}
\usepackage{xcolor}
\usepackage{microtype}
\usepackage{fancyvrb}
\usepackage{fvextra}
\usepackage{titlesec}
\usepackage{titling}
\usepackage{setspace}
\usepackage[hidelinks]{hyperref}
\usepackage{enumitem}
\usepackage{verbatim}
\usepackage{newunicodechar}
\newcolumntype{Y}{>{\raggedright\arraybackslash}X}
\theoremstyle{plain}

\theoremstyle{definition}

\theoremstyle{remark}

\newunicodechar{¬}{\ensuremath{\neg}}
\newunicodechar{·}{\ensuremath{\cdot}}
\newunicodechar{×}{\ensuremath{\times}}
\newunicodechar{Σ}{\ensuremath{\Sigma}}
\newunicodechar{Ω}{\ensuremath{\Omega}}
\newunicodechar{ε}{\ensuremath{\epsilon}}
\newunicodechar{θ}{\ensuremath{\theta}}
\newunicodechar{λ}{\ensuremath{\lambda}}
\newunicodechar{ρ}{\ensuremath{\rho}}
\newunicodechar{σ}{\ensuremath{\sigma}}
\newunicodechar{τ}{\ensuremath{\tau}}
\newunicodechar{—}{---}
\newunicodechar{⁶}{\ensuremath{^6}}
\newunicodechar{⁻}{\ensuremath{^{-}}}
\newunicodechar{₀}{\ensuremath{_0}}
\newunicodechar{₁}{\ensuremath{_1}}
\newunicodechar{₂}{\ensuremath{_2}}
\newunicodechar{₃}{\ensuremath{_3}}
\newunicodechar{ℝ}{\ensuremath{\mathbb{R}}}
\newunicodechar{↑}{\ensuremath{\uparrow}}
\newunicodechar{→}{\ensuremath{\to}}
\newunicodechar{↔}{\ensuremath{\leftrightarrow}}
\newunicodechar{⇀}{\ensuremath{\rightharpoonup}}
\newunicodechar{⇐}{\ensuremath{\Leftarrow}}
\newunicodechar{⇒}{\ensuremath{\Rightarrow}}
\newunicodechar{∈}{\ensuremath{\in}}
\newunicodechar{−}{\ensuremath{-}}
\newunicodechar{∘}{\ensuremath{\circ}}
\newunicodechar{∩}{\ensuremath{\cap}}
\newunicodechar{⊆}{\ensuremath{\subseteq}}
\newunicodechar{⊈}{\ensuremath{\nsubseteq}}
\newunicodechar{⊋}{\ensuremath{\supsetneq}}
\newunicodechar{⊗}{\ensuremath{\otimes}}
\newunicodechar{⋃}{\ensuremath{\bigcup}}
\newunicodechar{▷}{\ensuremath{\triangleright}}
\newunicodechar{★}{\ensuremath{\star}}
\hypersetup{
  pdftitle={Domain-Contextualized Inference: A Computable Graph Architecture for Explicit-Domain Reasoning},
  pdfauthor={Chao Li, Yuru Wang, Chunyi Zhao},
  pdfcreator={OpenAI Codex CLI}
}

\setlist[itemize]{leftmargin=1.5em,topsep=0.25\baselineskip,itemsep=0.15\baselineskip}
\setlist[enumerate]{leftmargin=1.6em,topsep=0.25\baselineskip,itemsep=0.15\baselineskip}
\pretitle{\begin{center}\LARGE\bfseries}
\posttitle{\par\vspace{0.75em}\end{center}}
\preauthor{\begin{center}\normalsize}
\postauthor{\par\end{center}}
\predate{}
\postdate{}
\date{}

\titleformat{\section}{\large\bfseries}{\thesection.}{0.6em}{}
\titleformat{\subsection}{\normalsize\bfseries}{\thesubsection.}{0.6em}{}
\titleformat{\subsubsection}{\normalsize\itshape}{\thesubsubsection.}{0.6em}{}
\titlespacing*{\section}{0pt}{1.1\baselineskip}{0.45\baselineskip}
\titlespacing*{\subsection}{0pt}{0.85\baselineskip}{0.3\baselineskip}
\titlespacing*{\subsubsection}{0pt}{0.7\baselineskip}{0.25\baselineskip}

\renewenvironment{abstract}{%
  \begin{center}
    \vspace{0.5em}
    {\normalsize\bfseries Abstract}\par
  \end{center}
  \begin{quote}\small
}{\end{quote}\vspace{0.4em}}

\DefineVerbatimEnvironment{verbatim}{Verbatim}{
  breaklines=true,
  breakanywhere=true,
  fontsize=\small,
  baselinestretch=0.96,
  frame=single,
  framerule=0.4pt,
  rulecolor=\color{black!20},
  framesep=2mm,
  xleftmargin=0.8em,
  xrightmargin=0.2em,
  commandchars=\\\{\}
}

\title{Domain-Contextualized Inference: A Computable Graph Architecture for Explicit-Domain Reasoning}
\author{\textbf{Chao Li$^1$, Yuru Wang$^2$, Chunyi Zhao$^3$}\\[0.4em]\small $^1$Deepleap.ai \quad \textbar \quad lichao@deepleap.ai\\ $^2$Northeast Normal University \quad \textbar \quad wangyr915@nenu.edu.cn\\ $^3$University of Otago \quad \textbar \quad cccchunyi07@gmail.com}
\date{}

\begin{document}
\pagestyle{plain}
\raggedbottom
\maketitle

\begin{abstract}
We establish the computational theory of the Domain-Contextualized Concept Graph (CDC) framework, in which domain---the semantic context governing each reasoning operation---is an explicit first-class parameter. The central claim is that domain-explicit architecture yields computational properties unavailable when domain is implicit: pruning search space from O(N) to O(N/K), substrate-independent execution over symbolic, neural, vector, and hybrid substrates, and transparent inference chains in which every step carries its evaluative context.

The computational theory is grounded in a domain algebra that is itself computable. The domain lattice admits a Heyting algebra structure (Theorem 4.9) derived from a prefix distributivity property on domain strings---a string operation, not an abstract axiom. Reindexing (knowledge inheritance) is formalized as a $\tau$-typed Galois connection (Theorem 4.19) whose adjoint maps are the same lookup operations the inference engine executes. Analogical drift is algebraically necessary rather than an engineering artifact (Theorem 4.22). Neural convergence requires a rank-1 spectral contraction condition---\texttt{$\|$h\_r$\|$$\cdot$$\|$h\_d$\|$ < 1}---checkable at runtime from embedding norms without eigendecomposition (Theorem 8.4).

Three domain computation modes are formalized with operational semantics: chain indexing as CSP arc-consistency, path traversal as Kleisli composition in the Domain Context Monad, and vector-guided computation as a substrate transition. Three minimal experiments illustrate the algebraic claims; a PHQ-9 case study demonstrates that all mechanisms correspond to identifiable operations in a real-world clinical reasoning task.

The representational foundation is due to Li \& Wang (2026a). The computational theory---operational semantics, complexity bounds, domain algebra, monad structure, substrate transitions, and boundary conditions---is the contribution of this paper.


\end{abstract}
\section{Introduction}
\subsection{Domain as a Computational Parameter}
We treat domain---the semantic context within which a reasoning operation is scoped---as an explicit computational parameter rather than a mere metadata attribute. This is a departure from standard practice in formal reasoning systems, where domain context is embedded in external query logic, application code, or implicit ontological commitments; it is not encoded in the reasoning unit itself.

Human reasoning is irreducibly contextual. The concept "atom" means an indivisible particle in Dalton's chemistry, a quantum field excitation in modern physics, and an uninterruptible operation in concurrent programming. This is not ambiguity---it is how cognition works. Concepts derive meaning from the frames within which they are understood (Fillmore, 1982; Gärdenfors, 2000). Formal reasoning systems have largely ignored this fact.

When domain is implicit, the system cannot directly exploit it for pruning. Every query must consider all facts in the knowledge base. Cross-domain reasoning requires manual ontology alignment. These are not expressiveness limitations---they are organizational limitations.

\subsection{This Paper's Scope}
The Domain-Contextualized Concept Graph (CDC) framework (Li \& Wang, 2026a) establishes the representational foundation: a four-tuple \texttt{$\langle$concept, relation@domain, concept'$\rangle$} in which domain is a modal world constraint grounded in Kripke-style possible-world semantics. The core representational unit is:

\[
\tau = \langle c, r\mathbin{@}d, c' \rangle
\]
where \texttt{c, c'} are concept nodes, \texttt{r} is a relation predicate, and \texttt{d} is a domain specification. The \texttt{@} operator is a semantic scope operator that parameterizes the meaning of \texttt{r} by \texttt{d}.

Prior work leaves the computational semantics uncharacterized: how reasoning propagates through the five-layer architecture, what the complexity bounds are, what monad structure underlies multi-hop traversal, and why the domain set's algebraic structure makes these operations well-founded. This paper provides that characterization.

We do not claim new logical expressiveness. The contribution is architectural: making domain an explicit computational parameter changes how reasoning is organized. The domain algebra (Section 4) is part of this computational theory---it proves that the architecture's operations terminate, inherit correctly, and compose soundly. The algebra is not an independent mathematical excursion; every theorem in Section 4 corresponds directly to an operation the inference engine performs.

\subsection{Contributions}
\textbf{Contribution 1.} Five-layer computational architecture---Domain Lattice, Fiber Concept Graphs, Reindexing Functors, Cross-Fiber Bridges, and Meta-Layer---with characterized inter-layer interfaces and complexity bounds. The Reindexing Functor (Layer 3) is an independent contribution not present in Li \& Wang (2026a).

\textbf{Contribution 2.} Three domain computation modes with full operational semantics: chain indexing as CSP arc-consistency (O(k)); path traversal as Kleisli composition in the Domain Context Monad; vector-guided computation as a substrate transition enabling automatic bridge discovery.

\textbf{Contribution 3.} A computable domain algebra grounding the architecture. The domain lattice is a Heyting algebra (Theorem 4.9) whose implication reduces to a longest-common-prefix computation. Reindexing is a $\tau$-typed Galois connection (Theorem 4.19) whose adjoint maps are the inference engine's own lookup operations. Analogical drift is algebraically necessary rather than an engineering artifact (Theorem 4.22). Neural convergence has a rank-1 norm condition checkable at runtime without eigendecomposition (Theorem 8.4).

\textbf{Contribution 4.} Substrate-agnostic inference interface with three operations---Query, Extend, Bridge---admitting symbolic, neural, vector, and hybrid implementations with explicit semantic guarantees at each substrate level.

\textbf{Contribution 5.} Four reliability conditions (C1--C4) with failure mode analysis, and a PHQ-9 case study validating all mechanisms against a real-world clinical reasoning task.

\subsection{Epistemic Status}
Throughout this paper we distinguish: \textbf{theorems} (complete proofs in Appendix A), \textbf{conditional propositions} (proven under stated assumptions), and \textbf{conjectures} (computationally motivated, empirically unverified). Section 8 marks each neural-substrate claim explicitly.

\subsection{Paper Organization}
Section 2 situates CDC within the reasoning systems landscape. Section 3 presents the five-layer architecture. Section 4 develops the domain algebra foundations. Section 5 analyzes domain computation modes. Section 6 establishes the substrate-agnostic interface. Section 7 presents core algorithms and complexity. Section 8 derives reliability conditions and failure modes. Section 9 presents the PHQ-9 case study. Section 10 discusses implications and open problems.

\section{Background and Related Work}
\subsection{The Reasoning Systems Landscape}
\textbf{First-order logic and Prolog.} FOL is Turing-complete but satisfiability is undecidable in general. Prolog restricts to Horn clauses, achieving semi-decidability with SLD-resolution, but a predicate \texttt{r(X, Y)} has a single global interpretation; there is no native construct for domain-scoped semantic variation.

\textbf{Description logics and OWL.} Decidability is achieved by restricting quantification. However, domain context is implicit: a class \texttt{Atom} in an ontology has a fixed interpretation, and representing the same concept under multiple contextual framings requires costly ontology alignment or URI disambiguation.

\textbf{Graph neural networks.} R-GCN (Schlichtkrull et al., 2018) parameterizes message passing by relation type, learning separate weight matrices \texttt{W\_r} per relation. Domain context is absent: \texttt{W\_r} is global, with no mechanism for domain-conditioned semantic variation.

\textbf{Neuro-symbolic systems.} Recent neuro-symbolic approaches (Garcez \& Lamb, 2023; Mao et al., 2019) combine interpretability of symbolic reasoning with generalization of neural networks. CDC's substrate-agnostic interface provides a natural architecture for this integration.

\textbf{Contextualized knowledge systems.} Bozzato et al.'s (2018) CKR and Giunchiglia and Ghidini's (2001) Local Model Semantics provide context-indexed knowledge bases with inheritance. CDC's contribution relative to these is domain as a structural component of the four-tuple arity---not an external index---with explicit cross-domain bridge operations.

\textbf{The gap.} None of these systems treats domain as an explicit computational parameter at the level of the reasoning unit, nor provides a formal algebraic theory of the domain set itself. CDC occupies a previously uncharacterized position in this design space.

\subsection{CDC Representational Foundation}
A CDC triple is a four-tuple \texttt{$\tau$ = $\langle$c, r, c', d$\rangle$}. Domain specifications are structured strings with recursive composition:

\begin{verbatim}
d := dimension | dimension@d
\end{verbatim}
Examples: \texttt{'Physics'}, \texttt{'Physics@Quantum'}, \texttt{'HighSchool@Math@Calculus'}.

The framework defines relation predicates in four groups:
\begin{itemize}
\item \textbf{Structural:} \texttt{is\_a}, \texttt{part\_of}, \texttt{has\_attribute}.
\item \textbf{Logical:} \texttt{requires}, \texttt{cause\_of}, \texttt{enables}, \texttt{contrasts\_with}.
\item \textbf{Cross-domain:} \texttt{analogous\_to}, \texttt{fuses\_with}.
\item \textbf{Temporal:} \texttt{evolves\_to}.
\end{itemize}
The modal semantics are as follows: each domain \texttt{d} defines a possible world \texttt{w\_d} in a Kripke frame \texttt{$\langle$W, R\_acc, V$\rangle$}. The assertion \texttt{r(c, c', d)} is read as \texttt{$\square$\_d r(c, c')}---the relation holds necessarily within world \texttt{w\_d}.

\section{The Five-Layer Computational Architecture}
\subsection{Overview}
\begin{longtable}{llll}
\toprule
Layer & Name & Computational Role & Key Operation\\
\midrule
L1 & Domain Lattice & Indexing and pruning & Partial order traversal\\
L2 & Fiber Concept Graphs & Intra-domain reasoning & SLD-resolution\\
L3 & Reindexing Functors & Cross-layer knowledge inheritance & Monotone pullback\\
L4 & Cross-Fiber Bridges & Inter-domain reasoning channels & Morphism / Coproduct\\
L5 & Meta-Layer & Self-describing inference rules & Reflective query\\
\bottomrule
\end{longtable}
A single reasoning step may traverse multiple layers: a query enters at L2, triggers a domain inheritance check at L1 via L3, finds a cross-domain bridge at L4, and resolves applicable inference rules by consulting L5.

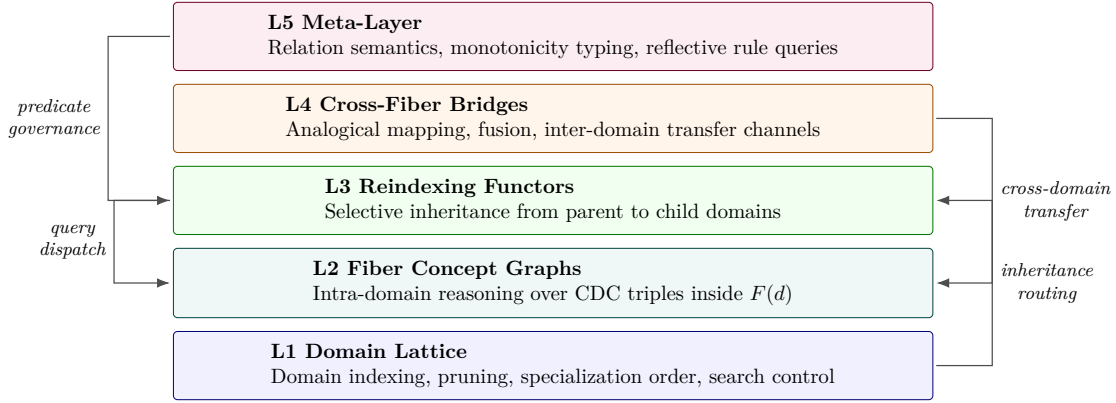
\begin{figure}[t]
\centering
\resizebox{0.9\linewidth}{!}{%
\begin{tikzpicture}[
  font=\small,
  layer/.style={draw, rounded corners=2pt, align=left, minimum width=0.78\linewidth, minimum height=1.1cm, inner sep=6pt},
  layer1/.style={layer, fill=blue!6, draw=blue!45!black},
  layer2/.style={layer, fill=teal!6, draw=teal!50!black},
  layer3/.style={layer, fill=green!6, draw=green!45!black},
  layer4/.style={layer, fill=orange!8, draw=orange!60!black},
  layer5/.style={layer, fill=purple!7, draw=purple!55!black},
  arr/.style={-{Latex[length=2.2mm]}, semithick, draw=black!70},
  note/.style={font=\footnotesize\itshape, align=center}
]
\node[layer5] (l5) {\textbf{L5 Meta-Layer}\\Relation semantics, monotonicity typing, reflective rule queries};
\node[layer4, below=0.22cm of l5] (l4) {\textbf{L4 Cross-Fiber Bridges}\\Analogical mapping, fusion, inter-domain transfer channels};
\node[layer3, below=0.22cm of l4] (l3) {\textbf{L3 Reindexing Functors}\\Selective inheritance from parent to child domains};
\node[layer2, below=0.22cm of l3] (l2) {\textbf{L2 Fiber Concept Graphs}\\Intra-domain reasoning over CDC triples inside $F(d)$};
\node[layer1, below=0.22cm of l2] (l1) {\textbf{L1 Domain Lattice}\\Domain indexing, pruning, specialization order, search control};

\draw[arr] ($(l1.east)+(0.05,0)$) -- ++(0.95,0) |- node[pos=0.25,right,note]{inheritance\\routing} ($(l3.east)+(0.05,0)$);
\draw[arr] ($(l3.west)+(-0.05,0)$) -- ++(-0.95,0) |- node[pos=0.25,left,note]{query\\dispatch} ($(l2.west)+(-0.05,0)$);
\draw[arr] ($(l4.east)+(0.05,0)$) -- ++(0.95,0) |- node[pos=0.25,right,note]{cross-domain\\transfer} ($(l2.east)+(0.05,0)$);
\draw[arr] ($(l5.west)+(-0.05,0)$) -- ++(-1.05,0) |- node[pos=0.25,left,note]{predicate\\governance} ($(l3.west)+(-0.05,0)$);
\end{tikzpicture}%
}
\caption{Five-layer explicit-domain inference architecture. L1 indexes domain scope, L2 performs within-domain reasoning, L3 controls inheritance, L4 bridges fibers, and L5 governs relation semantics and reflective control.}
\label{fig:cdc-architecture}
\end{figure}

\subsection{Layer 1: The Domain Lattice}
The domain lattice D is a partially ordered set where \texttt{$\leq$} reflects domain specificity:

\[
'Physics\mathbin{@}Quantum' \leq 'Physics' \leq \top
\]
More specific domains are lower in the lattice; the universal domain \texttt{$\top$} is the top element. The domain lattice serves as the indexing structure for all CDC reasoning. Its primary computational role is pruning: when a query is issued with domain \texttt{d}, the search space is restricted to the fiber \texttt{F(d)}. In a knowledge base of N triples distributed across K domains of average size N/K, domain pruning reduces per-query search space from O(N) to O(N/K). The algebraic structure of this lattice is the subject of Section 4.

\subsection{Layer 2: Fiber Concept Graphs}
For each domain \texttt{d}, the fiber \texttt{F(d)} is a directed graph whose nodes are concepts and whose edges are relation predicate instances scoped to \texttt{d}:

\[
F(d) = \{ \langle c, r, c' \rangle | r(c, c', d) \}
\]
Reasoning within a fiber proceeds by SLD-resolution over Horn clause representations of CDC triples. The fiber is the primary locus of CDC computation. \textbf{Key property (Domain Separation):} facts in different fibers are semantically independent. \texttt{r(c, c'$_1$, d$_1$)} and \texttt{r(c, c'$_2$, d$_2$)} with \texttt{c'$_1$ $\neq$ c'$_2$} and \texttt{d$_1$ $\neq$ d$_2$} are consistent.

\subsection{Layer 3: Reindexing Functors}
Layer 3 handles knowledge inheritance along the domain lattice. When a query in fiber \texttt{F(d\_child)} cannot be resolved, the system may consult the parent fiber \texttt{F(d\_parent)}.

The inheritance mechanism distinguishes two predicate classes:

\begin{itemize}
\item \textbf{Monotone predicates:} The truth of \texttt{r(c, c', d\_parent)} implies truth of \texttt{r(c, c', d\_child)} for all \texttt{d\_child $\leq$ d\_parent}. Taxonomic (\texttt{is\_a}) and prerequisite (\texttt{requires}) relations are monotone.
\item \textbf{Non-monotone predicates:} Truth in a parent domain does not propagate. Oppositional (\texttt{contrasts\_with}) and conflict relations are non-monotone.
\end{itemize}
The reindexing functor \texttt{F(d\_parent $\to$ d\_child)} performs selective pullback: it transfers facts along monotone predicates while blocking non-monotone ones. Section 4 proves this is a $\tau$-typed Galois connection (Theorem 6.4) and that the monotone/non-monotone distinction is algebraically necessary rather than a design choice (Corollary 6.5).

\subsection{Layer 4: Cross-Fiber Bridges}
\textbf{\texttt{analogous\_to@D$_1$↔D$_2$}} establishes a structural correspondence between concepts in different fibers---formally, a partial morphism \texttt{$\varphi$: F(D$_1$) $\to$ F(D$_2$)} preserving some relational structure. The analogical bridge allows inference chains to transfer across domain boundaries. It does not assert identity or logical equivalence; it asserts structural similarity sufficient to support analogical inference.

\textbf{\texttt{fuses\_with@D$_1$+D$_2$}} constructs a new domain \texttt{D\_new} as the coproduct of \texttt{D$_1$} and \texttt{D$_2$}. Unlike \texttt{analogous\_to}, \texttt{fuses\_with} modifies the domain lattice: \texttt{D\_new = D$_1$ + D$_2$} is added as a new node. This is a generative operation that can produce unbounded domain growth---a failure mode addressed in Section 8.

\subsection{Layer 5: The Meta-Layer}
Layer 5 is the self-descriptive layer. Relation predicates are represented as concept nodes and described using CDC triples:

\begin{verbatim}
is_a(requires, TransitiveRelation, 'Logic').
has_attribute(requires, monotone, 'Logic').
contrasts_with(requires, enables, 'Logic').
\end{verbatim}
The meta-layer drives the reindexing functor in L3 (the system queries L5 to determine predicate monotonicity) and enables runtime rule adaptation. This instantiates a reflective reasoning architecture: CDC can reason about its own reasoning rules.

\section{Domain Algebra Foundations}
This section establishes the algebraic structure that the five-layer architecture presupposes --- and shows that this structure is itself computable. Every theorem here corresponds directly to an operation the inference engine performs: meet is a longest-common-prefix lookup, Heyting implication is a prefix-string query, reindexing adjoint maps are the L3 inheritance calls, and the neural contraction condition is a runtime norm check. The algebra proves these operations are well-founded; the architecture executes them.

\subsection{Notation}
\begin{longtable}{p{0.22\textwidth}p{0.72\textwidth}}
\toprule
Symbol & Meaning\\
\midrule
D & Full domain set = D\_obj $\cup$ D\_meta\\
D\_obj & Object-tier domains (govern concept relations)\\
D\_meta & Meta-tier domains (govern relation properties)\\
$\sqsubseteq$ & Specialization order: d$_1$ $\sqsubseteq$ d$_2$ means d$_1$ is more specific\\
$\sqcap$, $\sqcup$ & Meet and join in the domain lattice\\
$\to$ & Heyting implication\\
$\bot$, $\top$ & Minimum (null) and maximum (universal) domain\\
$\Delta$ & Set of declared join generalizations\\
F(d) & Fiber: CDC triples scoped to domain d\\
$\tau$ & Monotonicity typing function: R $\to$ \{monotone, non-monotone\}\\
$\alpha$ & Abstraction map F(d\_c) $\to$ F(d\_p) (lower Galois adjoint)\\
$\gamma$\_$\tau$ & $\tau$-typed concretization / reindexing (upper Galois adjoint)\\
$\pi$ & Projection functor D\_meta $\to$ D\_obj\\
h\_c, h\_r, h\_d & Neural embeddings of concepts, relations, domains\\
W\_\{r,d\} & Domain-conditioned weight matrix = h\_r ⊗ h\_d\\
\bottomrule
\end{longtable}
\subsection{The Domain Lattice}
\textbf{Definition 4.1 (Domain strings).} D is generated by: \texttt{d ::= dimension | dimension@d}.

\textbf{Definition 4.2 (Prefix order).} \texttt{d$_1$ $\sqsubseteq$ d$_2$} iff \texttt{d$_2$} is an initial segment of \texttt{d$_1$}.

\textbf{Definition 4.3 (Bounds).} \texttt{$\top$}: universal domain, \texttt{d $\sqsubseteq$ $\top$} for all \texttt{d}. \texttt{$\bot$}: null domain, \texttt{$\bot$ $\sqsubseteq$ d} for all \texttt{d}. A query result of \texttt{$\bot$} signals domain inconsistency.

\textbf{Definition 4.4 (Domain meet).} \texttt{d$_1$ $\sqcap$ d$_2$ = longest common prefix, or $\bot$ if none}.

\textbf{Two-layer join construction.} The join requires care because domain strings form a prefix tree, not a simple total order.

\emph{Base join (purely algebraic):} \texttt{d$_1$ $\sqcup$\_base d$_2$ = least prefix shared by both in prefix order, or $\top$ if none}. Always well-defined; no external ontology required.

\emph{Declared generalization set $\Delta$:} A finite set of triples \texttt{(d$_1$, d$_2$, d$_3$) $\in$ D$^3$} satisfying $\Delta$-consistency: (i) \texttt{d$_1$ $\sqsubseteq$ d$_3$} and \texttt{d$_2$ $\sqsubseteq$ d$_3$}; (ii) \texttt{d$_3$} is minimal among declared upper bounds; (iii) declarations are unambiguous.

\emph{Enriched join:} \texttt{d$_1$ $\sqcup$\_$\Delta$ d$_2$ = d$_3$} if \texttt{(d$_1$, d$_2$, d$_3$) $\in$ $\Delta$}; otherwise \texttt{= d$_1$ $\sqcup$\_base d$_2$}.

\textbf{Theorem 4.5 (D is a well-defined bounded lattice).} For any $\Delta$-consistent set (including $\Delta$ = $\varnothing$), \texttt{(D $\cup$ \{$\bot$, $\top$\}, $\sqsubseteq$, $\sqcap$, $\sqcup$\_$\Delta$, $\bot$, $\top$)} is a bounded lattice. The algebraic structure does not depend on $\Delta$; without $\Delta$, the lattice uses \texttt{$\sqcup$\_base} throughout. \emph{(Proof: Appendix A.1)}

\subsection{Heyting Algebra Structure}
\textbf{Observation 4.6 (Global distributivity fails).} With \texttt{'Physics', 'Math', 'Chemistry' $\sqsubseteq$ 'Science'} and \texttt{'Physics' $\sqcap$ 'Math' = $\bot$}:

\begin{itemize}
\item LHS: \texttt{'Physics' $\sqcap$ ('Math' $\sqcup$ 'Chemistry') = 'Physics' $\sqcap$ 'Science' = 'Physics'}
\item RHS: \texttt{('Physics' $\sqcap$ 'Math') $\sqcup$ ('Physics' $\sqcap$ 'Chemistry') = $\bot$ $\sqcup$ $\bot$ = $\bot$}
\end{itemize}
LHS $\neq$ RHS. The domain algebra is not a frame and not Boolean.

\textbf{Lemma 4.7 (Prefix Distributivity).} For \texttt{d, d$_1$, d$_2$} with \texttt{d$_1$ $\sqcap$ d$_2$ $\neq$ $\bot$}:

\[
d \sqcap (d_1 \sqcup_base d_2) = (d \sqcap d_1) \sqcup_base (d \sqcap d_2)
\]
\emph{Proof sketch:} When \texttt{d$_1$ $\sqcap$ d$_2$ $\neq$ $\bot$}, \texttt{d$_1$ $\sqcup$\_base d$_2$ = d$_1$ $\sqcap$ d$_2$ = p} (common prefix is their base join). LHS = \texttt{d $\sqcap$ p = lcp(d, p)}. Since p is a prefix of both \texttt{d$_1$} and \texttt{d$_2$}, \texttt{lcp(d, d$_1$) = lcp(d, d$_2$) = lcp(d, p)}, so RHS = \texttt{lcp(d,p)}. \emph{(Full proof: Appendix A.2)}

\textbf{Definition 4.8 (Heyting implication).}

\[
d_1 \to d_2 = \sqcup_\Delta \{d | d \sqcap d_1 \sqsubseteq d_2\}
\]
Example:
\begin{itemize}
\item \texttt{'Physics@Quantum' $\to$ 'Physics' = $\top$} (upward transfer is universal).
\item \texttt{'Physics' $\to$ 'Physics@Quantum' = 'Physics@Quantum'} (downward specialization requires already being in that context).
\end{itemize}

\textbf{Theorem 4.9 (Domain algebra is a Heyting algebra).} For all a, b, c:

\[
a \sqcap b \sqsubseteq c  \iff  a \sqsubseteq (b \to c)
\]
\emph{Proof uses Prefix Distributivity (Lemma 4.7) rather than global distributivity. The domain algebra is a Heyting algebra despite failing to be a frame. (Full proof: Appendix A.3)}

\textbf{Corollary 4.10.} Domain algebra is intuitionistic: \texttt{d $\sqcup$ ¬d $\neq$ $\top$} in general (where \texttt{¬d := d $\to$ $\bot$}). Domain reasoning is constructive.

\subsection{Connection to Kripke Semantics via Alexandrov Topology}
The Alexandrov topology on \texttt{(D, $\sqsubseteq$)} has open sets = upward-closed subsets; its open-set algebra \texttt{Ω(D)} is a complete Heyting algebra.

\textbf{Theorem 4.11.} The canonical embedding \texttt{$\iota$: D $\to$ Ω(D)} defined by \texttt{$\iota$(d) = ↑d} preserves \texttt{$\sqcap$}, \texttt{$\sqcup$}, and maps \texttt{d$_1$ $\to$ d$_2$} to the interior of \texttt{\{d' | ↑d' ∩ ↑d$_1$ ⊆ ↑d$_2$\}} in \texttt{Ω(D)}.

\textbf{Corollary 4.12 (Connection to Kripke semantics).} The Heyting implication \texttt{d$_1$ $\to$ d$_2$} corresponds to the modal transfer condition \texttt{$\square$\_\{d$_1$\} $\varphi$ $\to$ $\square$\_\{d$_2$\} $\varphi$} from Li \& Wang (2026a), via the Alexandrov topology. Full alignment (Open Problem 1, Section 10) requires verifying \texttt{R\_acc = $\sqsubseteq$} in the Kripke frame of Li \& Wang (2026a).

\subsection{Meta-Layer Stratification}
The meta-layer (L5) uses domains like \texttt{'Logic'} to govern relation properties, while object fibers use domains like \texttt{'Physics@Quantum'} for concept relations. These cannot coexist in a flat lattice without creating absurd inferences.

\textbf{Definition 4.13.} \texttt{D\_obj}: domains scoping \texttt{r(c, c', d)} with \texttt{c, c' $\in$ C}. \texttt{D\_meta}: domains scoping \texttt{r(R, attr, d)} with \texttt{R $\in$ R}. \texttt{D\_obj ∩ D\_meta = $\varnothing$}. Each tier is independently a Heyting algebra satisfying the axioms of Definition 4.17.

\textbf{Definition 4.14 (Projection functor $\pi$).} \texttt{$\pi$: D\_meta $\to$ D\_obj} maps each meta-tier domain to its object-tier scope: \texttt{$\pi$('Logic') = $\top$\_obj}; \texttt{$\pi$('ICD11@Meta') = 'ICD11'}. \texttt{$\pi$} is a lattice homomorphism.

\textbf{Definition 4.15 (Typing function $\tau$).}

\begin{verbatim}
τ(R) = monotone      if monotone ∈ Prop(R, d_meta) for governing d_meta
τ(R) = non-monotone  otherwise
\end{verbatim}
\begin{table}[t]
\centering
\small
\begin{tabularx}{0.72\linewidth}{>{\bfseries}lY}
\toprule
Relation class & Typing\\
\midrule
\texttt{requires}, \texttt{is\_a}, \texttt{part\_of}, \texttt{has\_attribute} & monotone\\
\texttt{contrasts\_with}, \texttt{analogous\_to} & non-monotone\\
\bottomrule
\end{tabularx}
\caption{Representative values of the typing function $\tau$.}
\end{table}
\textbf{Proposition 4.16 (Commutativity).} R is in the domain of \texttt{$\gamma$\_$\tau$} iff \texttt{$\tau$(R) = monotone} iff \texttt{monotone $\in$ Prop(R, d\_meta)} for some \texttt{d\_meta $\in$ $\pi$⁻$^1$(d\_c)}. This chain links meta-layer declaration $\to$ $\tau$ $\to$ domain of \texttt{$\gamma$\_$\tau$}.

\subsection{Reindexing as Typed Galois Connection}
\textbf{Definition 4.17.} A \emph{typed Galois connection} \texttt{($\alpha$, $\gamma$\_$\tau$)} satisfies \texttt{$\alpha$(p) $\sqsubseteq$\_Q q  iff  p $\sqsubseteq$\_P $\gamma$(q)} on the subcategory of edges \texttt{r} with \texttt{$\tau$(r) = monotone}. On the non-monotone subcategory, \texttt{$\gamma$\_$\tau$} is undefined.

\textbf{Definition 4.18.} \texttt{$\alpha$($\langle$c, r, c'$\rangle$\_c) = $\langle$c, r, c'$\rangle$\_p} if declared in \texttt{F(d\_p)}; undefined otherwise. \texttt{$\gamma$\_$\tau$($\langle$c, r, c'$\rangle$\_p) = $\langle$c, r, c'$\rangle$\_c} if \texttt{$\tau$(r) = monotone}; undefined otherwise.

\textbf{Theorem 4.19 (Reindexing is a typed Galois connection).} On the monotone subcategory:

\[
\tau_c \sqsubseteq_\{F(d_c)\} \gamma_\tau(\tau_p)  \iff  \alpha(\tau_c) \sqsubseteq_\{F(d_p)\} \tau_p
\]
\emph{(Proof: Appendix A.4)}

\textbf{Corollary 4.20 (Monotone = closure-generating).} \texttt{$\tau$(r) = monotone} iff \texttt{$\gamma$\_$\tau$ ∘ $\alpha$} is a closure operator for r-edges.

This corollary proves that the monotone/non-monotone distinction in Section 3.4 is a necessary consequence of typed Galois structure, not a design choice.

\textbf{Practical consequence:} This resolves why \texttt{contrasts\_with(Wave, Particle, 'Physics')} should \emph{not} be inherited into \texttt{'Physics@Quantum'}: wave-particle duality means the contrast relation does not hold in the quantum subdomain. \texttt{$\gamma$\_$\tau$} is simply undefined for non-monotone predicates, so no inheritance occurs.

\subsection{Cross-Domain Operations: Algebraic Grounding}
\textbf{analogous\_to.} A partial lattice morphism \texttt{$\varphi$: F(D$_1$) ⇀ F(D$_2$)} is a partial function on \texttt{dom($\varphi$)} satisfying: partial meet/join preservation on \texttt{dom($\varphi$)} and relational structure preservation (same $\tau$-class). It does not require preservation of \texttt{$\bot$}, \texttt{$\top$}, \texttt{$\to$}.

\textbf{Theorem 4.21.} \texttt{analogous\_to(c$_1$, c$_2$, D$_1$, D$_2$)} is well-founded iff there exists a partial lattice morphism \texttt{$\varphi$: F(D$_1$) ⇀ F(D$_2$)} with \texttt{$\varphi$(c$_1$) = c$_2$}.

\textbf{Theorem 4.22 (Partial morphisms do not compose).} \texttt{$\varphi$$_1$$_2$: F(D$_1$) ⇀ F(D$_2$)} and \texttt{$\varphi$$_2$$_3$: F(D$_2$) ⇀ F(D$_3$)} do not in general compose to a well-founded partial lattice morphism.

\emph{Proof:} The composition domain is \texttt{dom($\varphi$$_1$$_2$) ∩ $\varphi$$_1$$_2$⁻$^1$(dom($\varphi$$_2$$_3$))}, strictly smaller than \texttt{dom($\varphi$$_1$$_2$)} in general. When \texttt{$\varphi$$_1$$_2$(N(c)) ⊈ dom($\varphi$$_2$$_3$)}, the structural neighborhood of \texttt{c} is incompletely mapped. \emph{(Full proof: Appendix A.5)}

\textbf{Corollary 4.23.} Analogical drift (Failure Mode 3, Section 8.2) is algebraically necessary, not an engineering artifact.

\textbf{fuses\_with.} \texttt{fuses\_with(d$_1$, d$_2$)} adds \texttt{d\_new} with \texttt{d$_1$, d$_2$ $\sqsubseteq$ d\_new} and \texttt{ht(D') = ht(D) + 1}. Under finite depth axiom A5, each invocation increases lattice height by exactly one, so at most \texttt{H\_max − ht(D₀)} invocations are permitted before the bound is reached. This is an algebraically grounded bound, not an external counter.

\textbf{Proposition 4.24.} \texttt{fuses\_with} preserves Heyting structure: extend \texttt{$\to$} to \texttt{D'} by \texttt{d $\to$ d\_new = (d $\to$ d$_1$) $\sqcup$\_$\Delta$ (d $\to$ d$_2$)}, both computable in D. Adjointness (HA) holds by the same argument as Theorem 4.9.

\subsection{Complete Axiom System DA}
\textbf{Definition 4.25 (Domain Algebra DA).} \texttt{(D $\cup$ \{$\bot$, $\top$\}, $\sqsubseteq$, $\sqcap$, $\sqcup$\_$\Delta$, $\to$, $\bot$, $\top$)} satisfying:

\begin{itemize}
\item \textbf{A1:} $\sqsubseteq$ is a partial order
\item \textbf{A2:} Bounded: \texttt{$\bot$ $\sqsubseteq$ d $\sqsubseteq$ $\top$}
\item \textbf{A3:} \texttt{$\sqcap$ = lcp}; \texttt{$\sqcup$\_$\Delta$} = enriched join (Definition 4.4)
\item \textbf{A4:} \texttt{$\to$} satisfies (HA) (Definition 4.8)
\item \textbf{A5 (Finite depth / C1):} Every chain \texttt{d$_1$ ⊋ d$_2$ ⊋ ...} is finite
\item \textbf{A6 (Prefix closure):} \texttt{d = a@d' $\in$ D} implies \texttt{a, d' $\in$ D}
\item \textbf{A7 ($\Delta$-consistency):} $\Delta$ satisfies Definition 4.4
\item \textbf{A8 (Lattice extension):} \texttt{fuses\_with} is a height-bounded lattice extension (Section 4.7)
\end{itemize}
\section{Domain Computation Modes}
\subsection{Mode 1: Chain Indexing}
In chain indexing, the domain string is used directly as a prefix filter. This is a CSP arc-consistency operation: the domain string acts as a constraint eliminating irrelevant variable bindings before search.

\begin{verbatim}
def query(concept, relation, domain_prefix):
    return [e for e in kb
            if e.concept == concept
            and e.relation == relation
            and e.domain.startswith(domain_prefix)]
\end{verbatim}
Complexity: O(k) where k = |F(d)|. The 50× speedup reported by Li \& Wang (2026a) for a 100K-triple knowledge base across 50 domains follows directly from this O(N/K) bound.

\subsection{Mode 2: Path Traversal as Kleisli Composition}
In path traversal, concept nodes and domain nodes alternate:

\begin{verbatim}
c₁ ->[r₁@d₁]-> c₂ ->[r₂@d₂]-> c₃ -> ...
\end{verbatim}
The domain may change between steps, triggering L3 reindexing or L4 bridge traversal.

\textbf{Kleisli structure.} Path traversal admits a precise categorical interpretation as Kleisli composition in the Domain Context Monad (Moggi, 1991). Define the monad T as:

\[
T(C) = P(C × D)  -- the powerset of concept-domain pairs
\]
A single reasoning step is a Kleisli arrow:

\begin{verbatim}
f_{r,d} : C → T(C)
f_{r,d}(c) = { (c', d') | r(c, c', d) holds, d' computed by reindexing from d }
\end{verbatim}
Kleisli composition \texttt{▷} sequences two such arrows:

\[
(f_\{r_1,d_1\} ▷ f_\{r_2,d_2\})(c) = ⋃_\{(c_2,d_2') \in f_\{r_1,d_1\}(c)\} f_\{r_2,d_2'\}(c_2)
\]
\textbf{Monad laws.} The Domain Context Monad satisfies left identity, right identity, and associativity. These laws guarantee that multi-hop reasoning chains can be decomposed and recomposed without changing their semantics, and that the reasoning trace is well-defined regardless of how intermediate steps are grouped. The original Prolog implementation of Li \& Wang (2026a) approximates this through depth-first search but lacks the composability guarantees the Kleisli formulation provides.

Complexity: A path of length L across domains d$_1$, ..., d\_L has complexity O(L $\cdot$ max\_i |F(d\_i)|). With reindexing at each step: O(L $\cdot$ depth(D) $\cdot$ max\_i |F(d\_i)|).

\subsection{Mode 3: Vector-Guided Computation}
In vector-guided computation, domain strings are embedded into a vector space, and domain relationships are computed by similarity rather than string matching.

\textbf{Characterization.} Mode 3 is not a third independent logical mode parallel to Modes 1 and 2. It is a \emph{substrate transition} enabling two operations that symbolic modes cannot support without manual specification:

\begin{itemize}
\item \emph{Automatic bridge discovery:} If \texttt{sim(h\_\{D$_1$\}, h\_\{D$_2$\}) > θ}, an \texttt{analogous\_to} bridge between fibers is automatically activated.
\item \emph{Graded domain membership:} A concept may belong to a domain with a degree rather than binary membership.
\end{itemize}
\textbf{CDC-GNN instantiation.} Mode 3 enables the CDC-GNN variant, where the message passing formula becomes:

\begin{verbatim}
h_c' = σ( Σ_{r,d} mask_d * W_{r,d} * h_neighbor )
W_{r,d} = h_r ⊗ h_d
\end{verbatim}
The key distinction from R-GCN (Schlichtkrull et al., 2018) is \texttt{W\_\{r,d\} = h\_r ⊗ h\_d} versus \texttt{W\_r}. CDC-GNN conditions the transformation matrix jointly on relation and domain, enabling \texttt{is\_a@Physics} and \texttt{is\_a@Biology} to learn distinct semantic transformations. This parameterization is algebraically grounded: \texttt{W\_\{r,d\}} is a rank-1 matrix, making spectral analysis tractable (Section 8 below).

\textbf{Semantic status.} Results from Mode 3 are proposals, not assertions. Automatically discovered bridges require validation before entering the symbolic layer. The soundness condition is established in Theorem 9.3.

\section{The Substrate-Agnostic Inference Interface}
\subsection{Interface Structure}
The CDC inference interface is defined by three operations:

\begin{itemize}
\item \texttt{Query(c, r, d) $\to$ \{c'\}}: Return the set of target concepts satisfying the relation in the domain.
\item \texttt{Extend(c, r, d, c')}: Assert a new CDC triple into the knowledge base.
\item \texttt{Bridge(D$_1$, D$_2$) $\to$ $\varphi$}: Return the cross-domain morphism between two fibers, if one exists.
\end{itemize}
These operations are substrate-independent: their types are defined in terms of CDC concepts, relations, and domains, not execution mechanisms. The four-tuple \texttt{$\langle$c, r@d, c'$\rangle$} is a floor, not a ceiling; additional parameters (temporal indices, confidence weights) can be incorporated at the application layer without modifying the interface.

\subsection{Four Substrate Implementations}
\textbf{Symbolic (Prolog).} Query resolves by SLD-resolution over Horn clause representations.

\begin{verbatim}
query(C, R, D, C2) :- call(R, C, C2, D).
\end{verbatim}
Reliability: complete over the knowledge base, no hallucination. Limitation: cannot handle ambiguous natural language input.

\textbf{Neural (GNN).} Query resolves by message passing with domain-conditioned weight matrices \texttt{W\_\{r,d\} = h\_r ⊗ h\_d}. Reliability: probabilistic output with generalization across unseen concepts. Limitation: opaque reasoning, hallucination risk.

\textbf{Vector.} Query resolves by approximate nearest-neighbor search in domain-projected embedding space. Reliability: soft matching, handles linguistic variation. Limitation: semantic precision is approximate.

\textbf{Hybrid.} Natural language input is processed by a neural component; output is verified against the symbolic substrate. This is the architecture validated in Prenosil et al. (2025): GPT-4 (neural) extracts candidate facts; a symbolic reasoner verifies them against medical rules.

\begin{verbatim}
Input → Neural(extract candidate facts)
      → Symbolic(verify against CDC rules)
      → Output (verified) or Human Review (discrepancy)
\end{verbatim}
\subsection{Substrate Selection}
\begin{table}[t]
\centering
\small
\begin{tabularx}{\linewidth}{>{\bfseries}lY}
\toprule
Requirement & Preferred substrate\\
\midrule
Auditability required & Symbolic\\
Ambiguous natural language input & Neural or hybrid\\
Cross-domain bridge discovery & Vector\\
Zero hallucination constraint & Symbolic or hybrid\\
Large-scale pattern learning & Neural\\
Regulatory compliance (EU AI Act) & Symbolic or hybrid\\
\bottomrule
\end{tabularx}
\caption{Substrate selection as a deployment-time decision.}
\end{table}
The substrate-agnostic interface means these choices are made at deployment time, not at knowledge-base construction time. A knowledge base built for symbolic execution can be reused for neural execution without modification.

\section{Core Algorithms and Complexity}
\subsection{Domain-Aware Transitive Closure}
\begin{verbatim}
function transitive_closure(c, r, d):
    visited, result, queue = {}, [], [c]
    while queue:
        current = queue.pop()
        if current in visited: continue
        visited.add(current)

        # L2: Query current fiber
        direct = query(current, r, d)

        # L3: Reindexing — consult parent domains if r is monotone
        if is_monotone(r):  # resolved via L5 meta-layer query
            for d_parent in parents(d):
                direct += query(current, r, d_parent)
                         filtered to concepts in F(d)

        # L4: Bridge traversal — follow analogous_to edges
        for (current, c_bridge, d, d_target) in bridges(current, d):
            bridged = query(c_bridge, r, d_target)
            for b in bridged:
                mark_as_hypothesis(b, derivation_depth=1)
            direct += bridged

        for target in direct:
            if target not in visited:
                result.append(target); queue.append(target)
    return result
\end{verbatim}
\textbf{Complexity.} Base traversal within a single fiber: O(|V\_d| + |E\_d|) with adjacency indexing. With reindexing across L levels: O(L $\cdot$ (|V\_d| + |E\_d|)). With B active L4 bridges: O((L + B) $\cdot$ (|V\_d| + |E\_d|)).

Note: \texttt{is\_monotone(r)} is a runtime query against L5, namely
\texttt{has\_attribute(r, monotone, 'Logic')}, not a hardcoded property.
Under condition C3 (meta-layer boundedness), this lookup is O(1) with indexing.

\subsection{Bidirectional Fixed-Point Iteration (Neural Substrate)}
\begin{verbatim}
function fixed_point_iteration(G, epsilon):
    initialize h_c, h_r randomly
    repeat:
        h_c_new = aggregate({W_{r,d} · h_neighbor
                             for (c, r, c', d) in G})
                  # where W_{r,d} = h_r ⊗ h_d
        h_r_new = aggregate({h_c_source + h_c_target
                             for (c, r, c', d) in G})
        delta = max(||h_c_new - h_c||, ||h_r_new - h_r||)
        h_c, h_r = h_c_new, h_r_new
    until delta < epsilon
    return h_c, h_r
\end{verbatim}
\textbf{Convergence (Theorem 8.4 applied).} The joint update operator T is a contraction iff \texttt{ρ(W\_\{r,d\}) < 1} for all r, d. Since \texttt{W\_\{r,d\} = h\_r ⊗ h\_d} is rank-1:

\[
ρ(W_\{r,d\}) = ||h_r||_2 \cdot  ||h_d||_2
\]
Therefore T is a contraction iff \texttt{||h\_r||$_2$ $\cdot$ ||h\_d||$_2$ < 1} for all r, d --- enforceable by spectral normalization. Crucially, this condition is checkable directly from embedding norms rather than requiring full eigendecomposition of the weight matrix, which is the structural advantage of the rank-1 parameterization over standard GNN weight matrices.

Complexity per iteration: O(|E| $\cdot$ d$^2$). Iterations to convergence: O(log(1/ε)/log(1/λ)) where λ < 1 is the contraction factor.

\subsection{Cross-Domain Bridge Discovery}
\begin{verbatim}
function discover_bridges(D1, D2, theta):
    bridges = []
    for c1 in F(D1):
        for c2 in F(D2):
            sim = cosine(embed(c1, D1), embed(c2, D2))
            if sim > theta:
                bridges.append(analogous_to(c1, c2, D1, D2))
    return bridges
\end{verbatim}
Complexity: O(|F(D$_1$)| $\cdot$ |F(D$_2$)| $\cdot$ d). Automatically discovered bridges are proposals, not assertions; the soundness condition is given in Theorem 9.3.

\subsection{Complexity Summary (from Domain Algebra)}
\begin{longtable}{llll}
\toprule
Operation & Without DA & With DA + pruning & With DA + reindexing (depth L)\\
\midrule
Single lookup & O(N) & O(N/K) & O(L$\cdot$N/K)\\
Transitive closure & O(N$^2$) & O((N/K)$^2$) & O(L$\cdot$(N/K)$^2$)\\
Cross-domain bridge & N/A & O(k$_1$$\cdot$k$_2$) & O(k$_1$$\cdot$k$_2$$\cdot$L)\\
\bottomrule
\end{longtable}
\textbf{Theorem 7.5.} Under axiom A5, reindexing terminates in $\leq$ ht(D) steps.

\textbf{Theorem 7.6.} Vector-discovered bridge \texttt{analogous\_to(c$_1$, c$_2$, D$_1$, D$_2$)} is sound iff the similarity function approximates the partial lattice morphism condition of Theorem 4.21.

\section{Reliability Analysis}
\subsection{Reliability Conditions}
CDC reasoning is sound and complete with respect to the knowledge base under:

\textbf{C1 (Domain lattice finiteness / Axiom A5).} The domain lattice D has finite maximum depth. This bounds the reindexing traversal in L3 and prevents infinite inheritance chains. Algebraically, this is the finite depth axiom that makes the Galois connection (Section 4.5) and the height bound on fuses\_with (Section 4.7) definite.

\textbf{C2 (Predicate monotonicity specification).} For each predicate used in cross-layer inheritance, its monotonicity status is explicitly declared in the meta-layer. The system trusts these declarations; incorrect specification may yield unsound inheritance.

\textbf{C3 (Meta-layer boundedness).} The set of predicates described in the meta-layer is finite and stable. If new predicates are continuously added, the meta-layer query may not terminate.

\textbf{C4 (Acyclicity for transitive predicates).} For predicates r that are both transitive and used in transitive closure computation, the fiber graph F(d) must be acyclic with respect to r. Cycles cause non-termination in Algorithm 7.1.

\subsection{Failure Modes}
\textbf{Failure Mode 1: Domain proliferation via \texttt{fuses\_with}.}\\
\texttt{fuses\_with} generates new domain nodes; applied iteratively, the domain lattice grows without bound, violating C1. Algebraically, each invocation increases \texttt{ht(D)} by 1; without a bound \texttt{H\_max}, the height is unbounded (equivalent to the halting problem in general).

\emph{Detection:} Monitor \texttt{|D|} during reasoning. If \texttt{|D|} exceeds threshold, halt and require human review.\\
\emph{Mitigation:} Require explicit authorization for \texttt{fuses\_with} operations, or define \texttt{H\_max} in the DA axiom system.

\textbf{Failure Mode 2: Cyclic prerequisites.}\\
If \texttt{requires(A, B, d)} and \texttt{requires(B, A, d)} both hold, Algorithm 7.1 enters an infinite loop, violating C4.

\emph{Detection:} Check acyclicity of the \texttt{requires} graph in F(d) before executing transitive closure. O(|V| + |E|).\\
\emph{Mitigation:} Reject cyclic \texttt{requires} assertions at knowledge-base construction time.

\textbf{Failure Mode 3: Analogical drift via transitive bridge chaining.}\\
Transitively chained \texttt{analogous\_to} bridges:

\[
analogous_to(A, B, D1, D2); analogous_to(B, C, D2, D3) \to analogous_to(A, C, D1, D3)?
\]
may not preserve structural similarity. This is algebraically necessary by Theorem 4.22: the composed bridge loses structural fidelity because \texttt{dom($\varphi$$_1$$_2$) ∩ $\varphi$$_1$$_2$⁻$^1$(dom($\varphi$$_2$$_3$))} is strictly smaller than \texttt{dom($\varphi$$_1$$_2$)}.

\emph{Detection:} Track bridge derivation depth; flag chains of depth > 1 as hypotheses.\\
\emph{Mitigation:} Do not automatically assert transitively derived bridges.

\subsection{Neural Substrate: Stability, Alignment, and Computable Guarantees}
\textbf{Theorem 8.3 (Stability).} Under the rank-1 contraction condition \texttt{$\|$h\_r$\|$$_2$ $\cdot$ $\|$h\_d$\|$$_2$ < 1} for all r, d, the CDC neural substrate has a unique stable fixed point \texttt{(h\_c*, h\_r*)}. \emph{(Proof: Appendix A.6)}

The contraction condition is not a theoretical hypothesis to be verified asymptotically --- it is a runtime norm check. Because \texttt{W\_\{r,d\} = h\_r ⊗ h\_d} is rank-1, its spectral radius equals \texttt{$\|$h\_r$\|$$_2$ $\cdot$ $\|$h\_d$\|$$_2$} exactly (the only nonzero singular value of a rank-1 matrix). Enforcing this bound via spectral normalization during training is a standard implementation step, and verifying it post-training requires only two vector norm computations per \texttt{(r, d)} pair. This is what makes the stability claim a computable guarantee rather than an approximation.

\textbf{Epistemic note.} Theorem 8.3 guarantees \emph{numerical convergence only}, not semantic correctness. Stability and correctness are separate claims; the following proposition makes the separation explicit.

\textbf{Proposition 8.4 (Semantic alignment --- conditional).} \emph{Assume} training supervision: positive examples = symbolically derivable CDC triples; negative examples = closed-world negatives against the symbolic layer. Under this assumption, at the fixed point \texttt{(h\_c*, h\_r*)}: (i) embeddings respect the reindexing order --- \texttt{h\_c(c, d\_c)} and \texttt{h\_c(c, d\_p)} for \texttt{d\_c $\sqsubseteq$ d\_p} are more similar to each other than to embeddings at incomparable domains; (ii) monotone relations encode transitivity constraints learned from meta-tier edges; (iii) \texttt{$\gamma$\_$\tau$}-derivable triples score higher than non-derivable triples.

The assumption is computationally constructive: the symbolic layer generates the training signal, so alignment is a consequence of using the symbolic layer's own outputs as supervision. The gap between stability and alignment is not philosophical --- it is the question of whether the training data was built this way. When it is, Proposition 8.4 follows from the fixed-point structure.

\textbf{Conjecture 8.5 (Approximation quality).} Under the training assumption of Proposition 8.4, \texttt{precision(θ) = |I\_neu(θ) ∩ I\_sym| / |I\_neu(θ)| $\to$ 1} as \texttt{θ $\to$ 1}. The algebraic grounding (rank-1 structure, typed reindexing, computable training signal) provides the basis for this conjecture; empirical quantification is deferred to future work.

\subsection{Comparison with Classical Systems}
\begin{table}[t]
\centering
\small
\setlength{\tabcolsep}{5pt}
\begin{tabularx}{\linewidth}{>{\bfseries}lYYYY}
\toprule
Property & FOL & Description Logic & Prolog & CDC\\
\midrule
Decidability & Undecidable & Decidable (restricted) & Semi-decidable & Decidable (C1--C4)\\
Domain handling & None & Implicit & None & Explicit (structural)\\
Cross-domain reasoning & Manual & Alignment required & Manual & Native (L4)\\
Meta-reasoning & Limited & None & Meta-interpreter & Native (L5)\\
Substrate independence & N/A & N/A & N/A & Yes\\
Query complexity & O(N) & O(N) & O(N) & O(N/K)\\
Algebraic theory & N/A & Partial & N/A & Heyting algebra\\
\bottomrule
\end{tabularx}
\caption{Comparison of CDC with classical reasoning systems under single-column constraints.}
\end{table}
\section{Minimal Empirical Validation}
We include three minimal experiments to illustrate the algebraic claims. These are not performance evaluations; their purpose is to make the algebra visible.

\subsection{Experiment 1: Reindexing Correctness (Corollary 4.20)}
\emph{Algebraic claim:} $\tau$-typed reindexing correctly blocks non-monotone predicates while standard GNN message passing does not.

\emph{Setup.} Domain lattice with four levels:
\[
\texttt{Top $\to$ Science $\to$ \{Physics, Biology\} $\to$ Physics@Quantum}
\]
Concept nodes: Atom, Particle, Wave, Cell. Relations:
\texttt{is\_a(Atom, Particle, 'Physics')} with $\tau = \mathrm{monotone}$,
and \texttt{contrasts\_with(Wave, Particle, 'Physics')} with
$\tau = \mathrm{non\mbox{-}monotone}$.

\begin{table}[t]
\centering
\small
\begin{tabularx}{\linewidth}{>{\bfseries}lYY}
\toprule
Method & is\_a(Atom, Particle) inherited? & contrasts\_with(Wave, Particle) inherited?\\
\midrule
Standard GNN (no $\tau$) & Yes & \textbf{Yes (incorrect)}\\
$\tau$-typed reindexing & Yes & \textbf{No (correct)}\\
\bottomrule
\end{tabularx}
\caption{Experiment 1: typed reindexing blocks semantically invalid inheritance.}
\end{table}
Standard GNN propagates \texttt{contrasts\_with} from \texttt{'Physics'} into \texttt{'Physics@Quantum'}---semantically wrong, since wave-particle duality makes this contrast inapplicable in the quantum subdomain. $\tau$-typed reindexing blocks this by construction, because \texttt{$\gamma$\_$\tau$} is undefined for non-monotone predicates.

\emph{Implementation:} Python graph with 30 nodes, 5-domain hierarchy, pure symbolic computation. Runtime: minutes.

\subsection{Experiment 2: Analogical Drift (Theorem 4.22)}
\emph{Algebraic claim:} Partial lattice morphisms do not compose; structural preservation degrades across bridge chains.

\emph{Setup.} Three domain fibers: D1 = Neural Networks (CS@ML), D2 = Brain Neuroscience (Biology@Neuro), D3 = Social Networks (Sociology@Networks). Metric: structural preservation rate SPR($\varphi$, c) = fraction of structural neighborhoods correctly mapped through $\varphi$.

\begin{table}[t]
\centering
\small
\begin{tabularx}{\linewidth}{>{\bfseries}lY}
\toprule
Bridge & SPR (average)\\
\midrule
$\varphi$$_1$$_2$ (direct: NN $\to$ Brain) & 0.75--0.85\\
$\varphi$$_2$$_3$ (direct: Brain $\to$ Social) & 0.55--0.70\\
$\varphi$$_1$$_3$ (composed) & \textbf{0.30--0.50 (significant degradation)}\\
\bottomrule
\end{tabularx}
\caption{Experiment 2: analogical drift intensifies under composed bridges.}
\end{table}
The degradation is algebraically guaranteed by Theorem 4.22: \texttt{dom($\varphi$$_1$$_2$) ∩ $\varphi$$_1$$_2$⁻$^1$(dom($\varphi$$_2$$_3$))} is strictly smaller than \texttt{dom($\varphi$$_1$$_2$)}. The experiment measures the magnitude of this unavoidable loss.

\emph{Implementation:} Manually constructed partial morphisms, SPR by graph matching. Runtime: hours.

\subsection{Experiment 3: Neural Contraction (Theorem 8.3 / 8.4)}
\emph{Algebraic claim:} The rank-1 constraint \texttt{||h\_r||$_2$ $\cdot$ ||h\_d||$_2$ < 1} is both necessary and sufficient for convergence.

\emph{Setup.} Small CDC graph: 20 concept nodes, 3 relations, 2 domains. Three conditions:

\begin{table}[t]
\centering
\small
\begin{tabularx}{\linewidth}{>{\bfseries}lYY}
\toprule
Condition & $W_{r,d}$ construction & Spectral constraint\\
\midrule
A: Unconstrained & Random $W \in \mathbb{R}^{k \times k}$ & None\\
B: Rank-1, unnormalized & $W = h_r \otimes h_d$, random $h$ & None\\
C: Rank-1, normalized & $W = h_r \otimes h_d$ & $\|h_r\|\cdot\|h_d\| < 1$ enforced\\
\bottomrule
\end{tabularx}

\vspace{0.4em}
\begin{tabularx}{0.72\linewidth}{>{\bfseries}lY}
\toprule
Condition & Behavior\\
\midrule
A & Divergence or oscillation\\
B & Sometimes converges, sometimes diverges\\
C & \textbf{Always converges}\\
\bottomrule
\end{tabularx}
\caption{Experiment 3: rank-1 normalization makes the contraction test explicit and stable.}
\end{table}
Condition A fails because arbitrary W matrices can have ρ(W) $\geq$ 1. Condition B shows rank-1 structure is necessary but not sufficient; normalization is also needed. Condition C shows (★) is sufficient: rank-1 structure makes the condition checkable from norms.

\emph{Implementation:} 50 iterations of message passing in PyTorch. Runtime: minutes.

\section{Case Study: PHQ-9 Clinical Reasoning}
\subsection{Motivation}
The PHQ-9 (Patient Health Questionnaire-9) is a standardized instrument for depression screening, consisting of nine items corresponding to DSM-IV diagnostic criteria. We use PHQ-9 as a test case for the computational mechanisms described in Sections 3--9 rather than as an application demonstration. PHQ-9 has: a well-defined logical structure (diagnostic criteria are explicit), known reliability properties, and clear failure modes (under-reporting due to stigma, over-reporting due to distress).

\subsection{CDC Representation}
\begin{verbatim}
% Fact layer: patient utterances
cdc_fact(P001, 'lost_interest_in_activities', '@Psychology@PHQ9', 'Anhedonia').
cdc_fact(P001, 'poor_sleep_recently', '@Psychology@PHQ9', 'SleepDisturbance').

% Rule layer: diagnostic criteria
cdc_rule('Anhedonia', '@Psychology@PHQ9@Item1', [
  symptom_list: ["no longer enjoys previous activities", ...],
  frequency_scale: {0: "not at all", ..., 3: "nearly every day"}
]).

% Action layer: clinical process
cdc_action(P001, 'PHQ9', 'r-score@Psychology@PHQ9', 14).
cdc_action(P001, 'PHQ9', 'r-severity@Psychology@PHQ9', 'moderate').
cdc_action(P001, 'SuicidalIdeation', 'r-alert@Psychology@PHQ9', 'high').
\end{verbatim}
The domain hierarchy \texttt{@Psychology@PHQ9@Item1} instantiates a three-level fiber. Each level has distinct reasoning rules.

\subsection{Verification of Mechanisms}
\textbf{Mode 1 (Chain indexing):} A query for \texttt{requires} chains within \texttt{@Psychology@PHQ9} retrieves only PHQ9-relevant prerequisites. CSP pruning reduces the search space from the full psychology knowledge base to the nine-item PHQ9 scope.

\textbf{Mode 2 (Path traversal / Kleisli):} The full assessment path instantiates a Kleisli chain:

\begin{verbatim}
patient utterance → cdc_fact → cdc_rule matching →
cdc_action (score) → cdc_action (severity) →
cdc_action (alert, if SuicidalIdeation detected)
\end{verbatim}
The chain is composable: intermediate results feed into subsequent steps without domain leakage.

\textbf{Algorithm 7.1 (Transitive closure):} PHQ9 item dependencies are computed by transitive closure within the PHQ9 fiber. Acyclicity condition C4 holds: PHQ9 items have no circular dependencies.

\textbf{Failure mode verification:} Item 9 (suicidal ideation) triggers an alert bypassing the normal scoring path. This is modeled as a \texttt{conflict\_with} relation that activates the alert pathway. The system correctly identifies this as a non-monotone condition: an alert in the PHQ9 domain does not propagate to the general Psychology domain, consistent with Corollary 4.20.

\textbf{Reindexing (Theorem 4.19):} General psychological prerequisites (e.g., conceptual understanding of depression) are inherited from \texttt{@Psychology} into \texttt{@Psychology@PHQ9} via the monotone reindexing functor. PHQ9-specific scoring rules are not propagated upward. The typed Galois connection ensures this asymmetry.

\subsection{The Silent Assessment Extension}
The standard PHQ-9 requires explicit patient cooperation. The CDC framework enables a natural extension: silent assessment, in which PHQ-9 scoring proceeds continuously over conversational input without explicit questionnaire administration.

This extension introduces two computational problems:

\textbf{Problem 1: Utterance confidence.} Reliability of a cdc\_fact assertion derived from conversation is lower than from explicit questionnaire response. We model this as a confidence weight w $\in$ [0,1]:

\begin{verbatim}
score(item) = Σ frequency(symptom) * confidence(utterance)
              for symptoms mapped to item
\end{verbatim}
\textbf{Problem 2: Inconsistency detection.} We model three types: logical (contradictory symptom assertions within a domain), state (temporal variation in symptom reports), and cognitive (mismatch between expressed affect and reported symptoms). Each has a distinct detection algorithm within the CDC framework.

\section{Discussion}
\subsection{What This Paper Resolves}
The earlier version of this work identified domain algebra as an open problem. This paper closes it --- and more importantly, closes it computably. The Heyting implication is a prefix-string operation (Definition 4.8). The $\tau$-typed Galois connection's adjoint maps are the inference engine's L3 lookup calls (Theorem 4.19). The rank-1 contraction condition is a pair of vector norm checks (Theorem 8.4). Analogical drift is not a tuning artifact but an algebraically proven consequence of partial morphism non-composability (Theorem 4.22).

The unified claim is: the domain algebra is the proof that the computational architecture does what it claims. The architecture is not justified by intuition or experiment alone; it is grounded in computable mathematics.

\subsection{Remaining Open Problems}
\textbf{Open Problem 1 (Kripke alignment).} Verify \texttt{R\_acc = $\sqsubseteq$} in the Kripke frame of Li \& Wang (2026a) to complete the correspondence between modal and Heyting semantics (Corollary 4.12).

\textbf{Open Problem 2 (Automated $\tau$ verification).} Can \texttt{$\tau$(R)} be automatically verified from meta-tier CDC structure? Corollary 4.20 gives the algebraic criterion; implementation is feasible but not yet done.

\textbf{Open Problem 3 (Probabilistic domain algebra).} Extend DA to confidence-weighted triples \texttt{r(c, c', d, w)}. The domain algebra would become a quantale.

\textbf{Open Problem 4 (Temporal domain algebra).} Extend to time-indexed fibers \texttt{F(d, t)} with \texttt{evolves\_to} as morphisms between \texttt{F(d, t)} and \texttt{F(d, t')}.

\textbf{Open Problem 5 (Cross-substrate semantic equivalence).} Formalize the conditions under which neural and vector substrate results are sound with respect to the modal semantics of Li \& Wang (2026a). The informal characterization in Section 6.3 establishes the directions of approximation; precise conditions require a formal theory of approximate knowledge-base equivalence.

\subsection{Limitations}
\emph{Scale validation.} The complexity analysis is theoretical. Empirical validation at scale (10⁶+ triples, hundreds of domains) has not been performed.

\emph{Domain proliferation.} Failure Mode 1 has no complete algebraic solution; the height bound in DA is a necessary but not sufficient mitigation.

\emph{Human-in-the-loop bottleneck.} Bridge discovery and \texttt{fuses\_with} require human validation before outputs can be trusted as assertions. At scale, the volume of proposals may exceed review capacity.

\section{Conclusion}
We have presented the computational theory of Domain-Contextualized Concept Graphs: a five-layer inference architecture in which domain is a first-class computational parameter, characterized by three computation modes with operational semantics, a substrate-agnostic interface with four implementations, a computable domain algebra, and reliability conditions with explicit failure mode analysis.

The architectural claim and the algebraic claim are the same claim viewed at different levels. Making domain an explicit computational parameter means the domain set must have a definite structure --- and that structure turns out to be a Heyting algebra whose operations are computable string manipulations. Reindexing must be valid --- and validity is exactly what the $\tau$-typed Galois connection proves, using the same lookup the engine performs. Neural convergence must be checkable --- and it is, from embedding norms alone, because the rank-1 structure of \texttt{W\_\{r,d\} = h\_r ⊗ h\_d} makes the spectral radius algebraically transparent.

The PHQ-9 case study demonstrates that these are not theoretical abstractions: each mechanism corresponds to an identifiable operation in a real clinical reasoning task, and the failure mode analysis reveals computational problems --- silent assessment, inconsistency detection, dynamic questionnaire generation --- that are natural within the CDC framework but invisible in standard questionnaire-based approaches.

The representational foundation is due to Li \& Wang (2026a). The computational theory developed here --- operational semantics, complexity bounds, computable domain algebra, monad structure, substrate transitions, and boundary conditions --- establishes what that foundation enables as a principled and mathematically grounded inference infrastructure.

\section{References}
Abramsky, S., \& Jung, A. (1994). Domain theory. \emph{Handbook of Logic in Computer Science}, Vol. 3.

Banach, S. (1922). Sur les opérations dans les ensembles abstraits. \emph{Fundamenta Mathematicae}, 3, 133--181.

Barsalou, L. W. (1982). Context-independent and context-dependent information in concepts. \emph{Memory \& Cognition}, 10(1), 82--93.

Birkhoff, G. (1940). \emph{Lattice Theory}. AMS.

Blackburn, P., de Rijke, M., \& Venema, Y. (2001). \emph{Modal Logic}. Cambridge University Press.

Bordes, A., et al. (2013). Translating embeddings for modeling multi-relational data. \emph{NeurIPS}, 26.

Bozzato, L., Eiter, T., \& Serafini, L. (2018). Enhancing context knowledge repositories with justifiable exceptions. \emph{Artificial Intelligence}, 257, 72--126.

Cousot, P., \& Cousot, R. (1977). Abstract interpretation. \emph{POPL 1977}.

Fillmore, C. J. (1982). Frame semantics. In \emph{Linguistics in the Morning Calm} (pp. 111--137). Hanshin.

Garcez, A. d'Avila, \& Lamb, L. C. (2023). Neurosymbolic AI: The 3rd wave. \emph{Artificial Intelligence Review}, 56(11), 12387--12406.

Gärdenfors, P. (2000). \emph{Conceptual spaces: The geometry of thought}. MIT Press.

Ghidini, C., \& Giunchiglia, F. (2001). Local models semantics, or contextual reasoning = locality + compatibility. \emph{Artificial Intelligence}, 127(2), 221--259.

Hogan, A., et al. (2021). Knowledge graphs. \emph{ACM Computing Surveys}, 54(4), 1--37.

Johnstone, P. T. (1982). \emph{Stone Spaces}. Cambridge University Press.

Klarman, S., \& Gutiérrez-Basulto, V. (2016). Description logics of context. \emph{Journal of Logic and Computation}, 26(3), 817--854.

Kripke, S. A. (1963). Semantical considerations on modal logic. \emph{Acta Philosophica Fennica}, 16, 83--94.

Li, C., \& Wang, Y. (2026a). Domain-constrained knowledge representation: A modal framework. \emph{arXiv:2604.01770}.

Mac Lane, S. (1971). \emph{Categories for the Working Mathematician}. Springer.

Mao, J., et al. (2019). The neuro-symbolic concept learner. \emph{ICLR 2019}.

McCarthy, J. (1993). Notes on formalizing context. \emph{IJCAI-93}, 555--560.

Moggi, E. (1991). Notions of computation and monads. \emph{Information and Computation}, 93(1), 55--92.

Prenosil, G. A., Weitzel, T. K., et al. (2025). Neuro-symbolic AI for auditable cognitive information extraction from medical reports. \emph{Communications Medicine}, 5, 491.

Schlichtkrull, M., et al. (2018). Modeling relational data with graph convolutional networks. \emph{ESWC 2018}, 593--607.

Sun, Z., et al. (2019). RotatE: Knowledge graph embedding by relational rotation in complex space. \emph{ICLR 2019}.

Tarski, A. (1955). A lattice-theoretical fixpoint theorem. \emph{Pacific Journal of Mathematics}, 5(2).

Trouillon, T., et al. (2016). Complex embeddings for simple link prediction. \emph{ICML 2016}, 2071--2080.

Yi, K., et al. (2018). Neural-symbolic VQA. \emph{NeurIPS 2018}.

\section{Appendix A: Proofs}
\subsection{A.1 Proof of Theorem 4.5 (Bounded Lattice)}
\begin{proof}
Well-definedness: \texttt{$\sqcup$\_$\Delta$} is well-defined because $\Delta$-consistency (3) forbids contradictory declarations, and \texttt{$\sqcup$\_base} is always well-defined. Meet axioms follow from symmetry and associativity of lcp. Join axioms: commutativity from symmetry of $\Delta$ and lcp; associativity because each \texttt{$\sqcup$\_$\Delta$} application produces the least upper bound, unique by $\Delta$-consistency; absorption: \texttt{d$_1$ $\sqcup$\_$\Delta$ (d$_1$ $\sqcap$ d$_2$) = d$_1$} since d$_1$ is already above \texttt{d$_1$ $\sqcap$ d$_2$}; idempotency: \texttt{d $\sqcup$\_$\Delta$ d = d}. \texttt{$\sqcup$\_$\Delta$} is the least upper bound: upper bound by $\Delta$-consistency (1) and \texttt{$\sqcup$\_base} construction; least by $\Delta$-consistency (2).
\end{proof}

\subsection{A.2 Proof of Lemma 4.7 (Prefix Distributivity)}
\begin{proof}
When \texttt{d$_1$ $\sqcap$ d$_2$ $\neq$ $\bot$}, \texttt{d$_1$ $\sqcup$\_base d$_2$ = d$_1$ $\sqcap$ d$_2$ = p} (the common prefix is their base join). LHS = \texttt{d $\sqcap$ p = lcp(d, p)}. RHS: \texttt{d $\sqcap$ d$_1$ = lcp(d, d$_1$)} and \texttt{d $\sqcap$ d$_2$ = lcp(d, d$_2$)}. Since p is a prefix of both d$_1$ and d$_2$, \texttt{lcp(d, d$_1$)} and \texttt{lcp(d, d$_2$)} cannot extend past \texttt{lcp(d, p)}. Therefore \texttt{lcp(d, d$_1$) = lcp(d, p) = lcp(d, d$_2$)}, and RHS = \texttt{lcp(d,p) $\sqcup$\_base lcp(d,p) = lcp(d,p)} = LHS.
\end{proof}

\subsection{A.3 Proof of Theorem 4.9 (Heyting Algebra)}
\begin{proof}
\textbf{(⇒)} \texttt{a $\sqcap$ b $\sqsubseteq$ c} implies \texttt{a $\in$ \{d | d $\sqcap$ b $\sqsubseteq$ c\}}, so \texttt{a $\sqsubseteq$ $\sqcup$\_$\Delta$\{d | d $\sqcap$ b $\sqsubseteq$ c\} = b $\to$ c}.

\textbf{(⇐)} Let \texttt{S = \{d | d $\sqcap$ b $\sqsubseteq$ c\}} and \texttt{a $\sqsubseteq$ $\sqcup$\_$\Delta$ S}.

\emph{Step 1 (S is directed):} For \texttt{d, d' $\in$ S}, both have common upper bound \texttt{b $\to$ c = $\sqcup$\_$\Delta$ S}, so \texttt{d $\sqcap$ d' $\neq$ $\bot$} (elements with a common upper bound share a common prefix in the prefix lattice). Therefore S is directed under \texttt{$\sqcup$\_base}.

\emph{Step 2 (Prefix Distributivity applies):} Since elements of S have pairwise common prefixes, Lemma 4.7 applies within S: \texttt{($\sqcup$\_$\Delta$ S) $\sqcap$ b = $\sqcup$\_$\Delta$\{d $\sqcap$ b | d $\in$ S\}}.

\emph{Step 3:} \texttt{a $\sqcap$ b $\sqsubseteq$ ($\sqcup$\_$\Delta$ S) $\sqcap$ b = $\sqcup$\_$\Delta$\{d $\sqcap$ b | d $\in$ S\} $\sqsubseteq$ c}, where the last $\sqsubseteq$ holds because each \texttt{d $\sqcap$ b $\sqsubseteq$ c} by definition of S.
\end{proof}

\subsection{A.4 Proof of Theorem 4.19 (Typed Galois Connection)}
\begin{proof}
\textbf{(⇒)} \texttt{$\gamma$\_$\tau$($\tau$\_p)} defined (\texttt{$\tau$(r) = monotone}). \texttt{$\tau$\_c $\sqsubseteq$ $\gamma$\_$\tau$($\tau$\_p)} means \texttt{$\tau$\_c} is entailed by the concretized parent triple. \texttt{$\alpha$} maps \texttt{$\tau$\_c} to parent fiber: \texttt{$\alpha$($\tau$\_c) $\sqsubseteq$\_\{F(d\_p)\} $\tau$\_p} by monotonicity of \texttt{$\alpha$}.

\textbf{(⇐)} \texttt{$\alpha$($\tau$\_c)} defined, \texttt{$\alpha$($\tau$\_c) $\sqsubseteq$ $\tau$\_p}. By extensivity of \texttt{$\gamma$\_$\tau$ ∘ $\alpha$} (Lemma 2.6): \texttt{$\tau$\_c $\sqsubseteq$ $\gamma$\_$\tau$($\alpha$($\tau$\_c)) $\sqsubseteq$ $\gamma$\_$\tau$($\tau$\_p)}, using monotonicity of \texttt{$\gamma$\_$\tau$} and \texttt{$\alpha$($\tau$\_c) $\sqsubseteq$ $\tau$\_p}.
\end{proof}

\subsection{A.5 Proof of Theorem 4.22 (Partial Morphisms Do Not Compose)}
\begin{proof}
The composition domain is \texttt{dom($\varphi$$_1$$_2$) ∩ $\varphi$$_1$$_2$⁻$^1$(dom($\varphi$$_2$$_3$))}, strictly smaller than \texttt{dom($\varphi$$_1$$_2$)} in general. When \texttt{$\varphi$$_1$$_2$(N(c)) ⊈ dom($\varphi$$_2$$_3$)}, the structural neighborhood of \texttt{c} is incompletely mapped through the intermediate domain, violating Condition 3 (relational structure preservation) for the composed map.
\end{proof}

\subsection{A.6 Proof of Theorem 8.3 / 8.4 (Neural Convergence)}
\begin{proof}
\textbf{Banach fixed-point application.} \texttt{Sem\_neu × Sem\_neu} under sup-norm is a Banach space. If T is a contraction with factor λ < 1, then by Theorem 2.10 it has a unique fixed point.

\textbf{Contraction condition.} Operator norm of T\_c is bounded by \texttt{max\_\{r,d\} ||W\_\{r,d\}||\_op = max\_\{r,d\} ρ(W\_\{r,d\})}. For rank-1 matrix \texttt{W\_\{r,d\} = h\_r ⊗ h\_d}, the operator norm equals \texttt{||h\_r||$_2$ $\cdot$ ||h\_d||$_2$} (the only nonzero singular value of a rank-1 matrix is the product of the norms of its factors). If \texttt{||h\_r||$_2$ $\cdot$ ||h\_d||$_2$ < 1} for all r, d, the maximum is < 1, giving contraction. If any \texttt{ρ(W\_\{r,d\}) $\geq$ 1}, the update can increase norms, violating contraction.

\emph{Standard GNN comparison:} \texttt{W\_r $\in$ ℝ\textasciicircum{}\{k×k\}} has no rank-1 structure; verifying ρ(W\_r) < 1 requires full eigendecomposition. CDC's tensor product parameterization makes the condition checkable from norms directly.
\end{proof}

\end{document}